\documentclass{article}

\PassOptionsToPackage{numbers}{natbib}
\usepackage[final]{neurips_2024}

\usepackage[pdftex]{graphicx}
\usepackage{amsthm}
\usepackage[utf8]{inputenc} 
\usepackage[T1]{fontenc}    
\usepackage{hyperref}       
\usepackage{url}            
\usepackage{booktabs}       
\usepackage{amsfonts}       
\usepackage{nicefrac}       
\usepackage{microtype}      
\usepackage{xcolor}         
\usepackage{algorithm}
\usepackage{booktabs}
\usepackage{array}
\usepackage{makecell}
\usepackage{times}
\usepackage{subcaption}
\usepackage{caption}

\usepackage[utf8]{inputenc} 
\usepackage[T1]{fontenc}    
\usepackage{hyperref}       
\usepackage{url}            
\usepackage{booktabs}       
\usepackage{amsfonts,amsmath,amssymb}       
\usepackage{nicefrac}       
\usepackage{microtype}      
\usepackage{xcolor}         
\usepackage{txfonts}
\usepackage{graphicx}
\usepackage{wrapfig,bm,comment,color}
\usepackage{breakurl,epsfig,epsf,fmtcount,semtrans,multirow,boldline}
\usepackage{tcolorbox}
\tcbuselibrary{skins}
\usepackage{tikz}

\definecolor{darkred}{RGB}{150,0,0}
\definecolor{darkgreen}{RGB}{0,150,0}
\definecolor{darkblue}{RGB}{0,0,200}
\hypersetup{colorlinks=true, linkcolor=darkred, citecolor=darkgreen, urlcolor=darkblue}
\newcommand{\xuechen}[1]{\textcolor{black}{#1}}

\newtheorem{claim}{Claim}
\newtheorem{assumption}{Assumption}

\newtheorem{lemma}{Lemma}

\newtheorem{proposition}{Proposition}
\newtheorem{definition}{Definition}

\newtheorem{example}{Example}

\newcommand{\beq}{\begin{equation}}
\newcommand{\ba}{\begin{align}}
\newcommand{\ea}{\end{align}}

\newcommand{\eeq}{\end{equation}}
  
\newcommand{\vct}[1]{\bm{#1}}
\newcommand{\mtx}[1]{\bm{#1}}

\newcommand{\eps}{\varepsilon}
\newcommand{\map}{\texttt{map}}

\newcommand{\st}{\star}

\newcommand{\A}{{\mtx{A}}}

\newcommand{\V}{{\mtx{V}}}

\newcommand{\Lc}{{\cal{L}}}

\newcommand{\Nc}{{\cal{N}}}

\newcommand{\Dc}{{\cal{D}}}

\newcommand{\Pb}{{\mtx{P}}}

\newcommand{\Qb}{{\mtx{Q}}}
\newcommand{\Cb}{{\mtx{C}}}
\newcommand{\Eb}{{\mtx{E}}}

\newcommand{\Iden}{{\mtx{I}}}
\newcommand{\M}{{\mtx{M}}}
\newcommand{\z}{{\vct{z}}}

\newcommand{\Nn}{\mathcal{N}}
\newcommand{\vb}{\vct{v}}

\newcommand{\s}{\vct{s}}
\newcommand{\ab}{\vct{a}}
\newcommand{\bb}{\vct{b}}

\newcommand{\x}{\vct{x}}

\newcommand{\y}{\vct{y}}

\newcommand{\W}{\mtx{W}}

\newcommand{\Vc}{{\cal{V}}}
\newcommand{\Pc}{{\cal{P}}}
\newcommand{\X}{{\mtx{X}}}

\newcommand{\Vb}{{\mtx{V}}}

\newcommand{\Kb}{{\mtx{K}}}

\newcommand{\qb}{{\vct{q}}}
\newcommand{\eb}{\vct{e}}

\newcommand{\pow}{\texttt{pow}}

\newcommand{\R}{\mathbb{R}}

\newcommand{\bs}{{\bar{\vct{s}}}}
\newcommand{\csf}[2]{\mathbb{S}_{\leq #1}(#2)}

\newcommand{\E}{\operatorname{\mathbb{E}}}

\newcommand{\sft}[1]{\mathbb{S}(#1)}
\newcommand{\att}[1]{\texttt{att}(#1)}

\newcommand{\tn}[1]{\|{#1}\|_{2}}

\newcommand{\tin}[1]{\|{#1}\|_{\ell_\infty}}

\usepackage{cleveref}
\title{GSA}
\title{Selectivity Improves Attention Mechanism in Language Modeling}
\title{A Selective Attention Mechanism for Enhancing Context Control in Language Models}
\title{Attention with Token and Position Selectivity Enhances Context Control in Language Models}
\title{Enhancing Transformers through Token and Position Selectivity}
\title{Selective Attention: Enhancing Transformer through Principled Context Control}

\author{%
  Xuechen Zhang \\
  University of Michigan \\
  zxuechen@umich.edu \\
  \And
    Xiangyu Chang \\
  University of California, Riverside \\
  cxian008@ucr.edu \\
      \And 
    Mingchen Li \\
  University of Michigan \\
  milii@umich.edu \\
      \And  
          Amit Roy-Chowdhury \\
  University of California, Riverside \\
  amitrc@ece.ucr.edu \\
  \And
    Jiasi Chen \\
 University of Michigan \\
  jiasi@umich.edu \\
    \And  
  Samet Oymak \\
 University of Michigan \\
  oymak@umich.edu \\
    \And
}

\begin{document}

\maketitle

\begin{abstract}
The attention mechanism within the transformer architecture enables the model to weigh and combine tokens based on their relevance to the query. While self-attention has enjoyed major success, it notably treats all queries $q$ in the same way by applying the mapping $V^\top\text{softmax}(Kq)$, where $V,K$ are the value and key embeddings respectively. In this work, we argue that this uniform treatment hinders the ability to control contextual \xuechen{sparsity} and relevance. As a solution, we introduce the ``\emph{Selective Self-Attention}'' (SSA) layer that augments the softmax nonlinearity with a principled temperature scaling strategy. By controlling temperature, SSA adapts the contextual \xuechen{sparsity} of the attention map to the query embedding and its position in the context window. Through theory and experiments, we demonstrate that this alleviates attention dilution, aids the optimization process, and enhances the model's ability to control softmax spikiness of individual queries. We also incorporate temperature scaling for value embeddings and show that it boosts the model's ability to suppress irrelevant/noisy tokens. Notably, SSA is a lightweight method which introduces less than 0.5\% new parameters through a weight-sharing strategy and can be fine-tuned on existing LLMs. Extensive empirical evaluations demonstrate that SSA-equipped models achieve a noticeable and consistent accuracy improvement on language modeling benchmarks. 

\end{abstract}


\section{Introduction}\label{sec intro}

Attention is a pivotal mechanism in modern machine learning that allows the model to focus on and retrieve different parts of the data, enhancing its ability to capture contextual relationships across time and space. While it was originally developed for NLP tasks through the transformer architecture, it has enjoyed widespread success in other domains such as computer vision, sequence modeling, and reinforcement learning \cite{vaswani2017attention,raffel2020exploring,brown2020language,chowdhery2023palm,sanh2021multitask}.

The canonical self-attention mechanism is a sequence-to-sequence map that outputs $\X\rightarrow\sft{\Qb\Kb^\top}\Vb$ where $\sft{\cdot}$ denotes the row-wise softmax nonlinearity and $\Qb$, $\Kb$, $\Vb$ are the query, key, and value embeddings obtained through linear projections of the input sequence $\X$. Through this process, for each query, the model creates a query-dependent composition of the input context. Importantly, the model has to accomplish two objectives: namely, capturing \emph{semantic similarity} between tokens and also adjusting the \emph{contextual sparsity}. Here, semantic similarity can be quantified through the angle between key-query embeddings and the contextual \xuechen{sparsity} through the spikiness of the attention map. While the importance of the former is clear, the latter is equally important given the fact that attention maps tend to be sparse in practice \cite{child2019generating,tarzanagh2023maxmargin,sahiner2022unraveling,chen2021scatterbrain}.

\begin{figure}[htbp]
\centering
\vspace{-15pt}
\begin{subfigure}[b]{\textwidth}
    \centering
	\begin{tikzpicture}
		\node at (0,0) []{\includegraphics[width=0.9\textwidth]{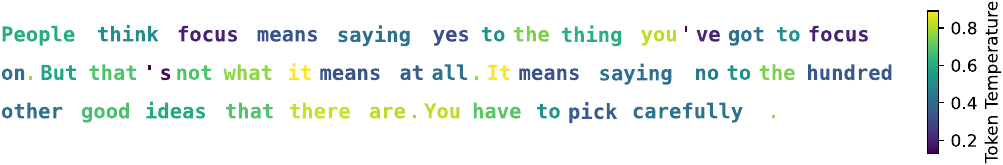}};
\end{tikzpicture}
\vspace{-6pt}
\vspace{-3pt}
\end{subfigure}
\caption{A quotation by Steve Jobs. We highlight tokens according to their temperatures learned by the SSA layer. Darker colors correspond to lower temperatures and receive a sparser attention map.}\label{fig:text_heat}
\vspace{-0.1cm}
\end{figure}

In this paper, we argue that these two objectives can be at odds and, as a result, the self-attention layer may struggle to achieve both objectives simultaneously due to its relatively inflexible parameterization. To address this issue, we propose the \emph{Selective Self-Attention} (SSA) layer that aims to \emph{decouple semantic similarity from contextual \xuechen{sparsity}}. SSA relies on a principled application of temperature-scaling (TS) to query and value embeddings. For instance, given query embedding $\qb$, rather than computing $\sft{\Kb\qb}$, SSA computes $\sft{\tau(\qb)\cdot\Kb\qb}$ where $\tau(\qb)$ is the learnable inverse-temperature. Intuitively, this allows for better control of the context window because $\tau(\qb)$ can control \emph{contextual \xuechen{sparsity}} while the projection matrices $\W_k,\W_q$ can fully focus on controlling semantic similarity. Figure \ref{fig:text_heat} shows an example of the learned token temperatures when training the Pythia model with SSA. In summary, we make the following theoretical and empirical contributions:


\begin{itemize} 
\item \textbf{Query selectivity.} We prove that introducing TS to the query embeddings enhances the model's capability to express a target attention map with smaller parameter norms (\Cref{lem soft approx}). TS particularly helps when attention maps exhibit large variations in \xuechen{spikiness} across different queries. Real and synthetic experiments corroborate that TS enables spikier/sharper attention maps and mitigates attention dilution. See Figure \ref{fig:diff} as an illustration.  
\item \textbf{Value selectivity.} We formalize the benefit of TS on value embeddings through a denoising perspective. Namely, we describe a denoising task where the linear value projection fails to filter the noisy tokens, and demonstrate how nonlinear scaling can boost denoising capability.
\item \textbf{Positional temperature.} We incorporate a term that adjusts the query-temperature according to the position in the context window. We show that this term can mitigate the dilution of attention scores caused by the increasing context length.


\item \textbf{Modularity and parameter-efficiency of SSA.} Selective Attention is accomplished by introducing a parameter-efficient temperature module that can be easily integrated into existing attention models. In practice, this introduces 5\% additional parameters to the model. We also introduce a weight sharing strategy that reduces the number of parameter overhead to less than 0.5\% while maintaining the benefits of SSA. We reuse the attention weights within the temperature module, which results in negligible inference/latency overhead since no additional matrix multiplication is required. These methods only involve vector dot-products (at the output layer of the temperature module) and elementwise scaling of matrices.

\item \textbf{Empirical benefits.} Our evaluations on the NLP benchmarks of Wikitext~\cite{merity2016pointer}, Lambada~\cite{paperno2016lambada}, Piqa~\cite{bisk2020piqa}, Hella~\cite{zellers2019hellaswag}, Winogrande~\cite{sakaguchi2021winogrande}, Arc-E, and Arc-C~\cite{clark2018think} demonstrate that Selective Attention noticeably improves language modeling performance. These benefits are consistent across various models including GPT-2~\cite{radford2019language}, Pythia~\cite{biderman2023pythia}, Llama~\cite{touvron2023llama} and Llama3~\cite{dubey2024llama}, as well as during both fine-tuning and pre-training, as shown in \Cref{tab:my_label}. Additionally, evaluations on the passkey retrieval task~\cite{peng2023yarn,mohtashami2023landmark} reveal that SSA substantially enhances the retrieval capabilities of the transformer, shown in \Cref{tab:passkey}.

\end{itemize}

\section{Related Work}


\noindent\textbf{Temperature Scaling (TS):} 
TS is a fundamental method for controlling model behavior, influencing aspects such as stochasticity of generative LLMs, calibration and uncertainty, and imbalanced data, as highlighted in several studies~\cite{menon2020long,li2021autobalance,zhang2024class}. Related to us, previous research \cite{peng2023yarn,yao-etal-2021-self,chi2023attention} has also proposed utilizing a temperature term in the softmax function to enhance the \emph{length extrapolation} capabilities of transformers. For instance, Yarn \cite{peng2023yarn} scales the attention logits as a function of the sequence length and shows that this improves the perplexity when extending the context window. Our work provides a formal justification for the temperature scaling rule proposed in Yarn (see Proposition \ref{position thm}) and also highlights the value of adapting temperature to the individual positions. Importantly, our approach is differentiable and obviates the need for grid search required by prior works. Since we don't focus on length generalization, we have found that position-aware temperature has a much smaller benefit compared to token-aware temperature, which is our primary contribution.

\noindent\textbf{Gating mechanisms and selectivity:} Various strategies have been developed to mitigate the impact of uninformative inputs in model training and processing.
Gating mechanisms, originally introduced through LSTMs \cite{hochreiter1997long}, have been proposed to selectively filter or scale down the input sequence \cite{yang2023gated,dauphin2017language,de2024griffin,mehta2023long,shazeer2020glu}. Very recent sequence models such as Mamba (a.k.a.~selective state-space model) and Griffin also incorporate gating to boost language modeling \cite{gu2023mamba,wang2024mambabyte,zhu2024vision,de2024griffin,katsch2023gateloop}. These models leverage input-dependent gating to ensure parallellizable training and enjoyed noticeable success. These methodologies inspired our approach, which incorporates TS to augment the selection capabilities of the attention layer. Specifically, TS can be viewed as an instance of gating that selectively passes or suppresses tokens to provide better control of contextual \xuechen{sparsity} and relevance. In this light, our work also provides a mechanistic understanding of how gating mechanism can aid self-attention to improve its expressive capabilities. Finally, we highlight the concurrent work \cite{ye2024differential} which utilizes a \emph{differential softmax parameterization} to promote spiky attention maps. 

\noindent\textbf{Mechanistic understanding of transformers:} The importance of transformer-based models led to many research efforts on developing a stronger understanding of various aspects of transformer and attention \cite{olsson2022context,xie2021explanation,dong2021attention}. While it is impossible to cover all of these works, it is evident that capability to select relevant features and promote contextual \xuechen{sparsity} is crucial for the ability of language models to perform complex tasks such as reasoning \cite{liu2024exposing,abbe2023generalization,tarzanagh2023maxmargin,zhou2023algorithms}. These have provided inspiration for us to pursue an enhanced modeling of attention's \xuechen{spikiness} (e.g.~as in Figure \ref{fig:diff}). The experiments in Figure \ref{fig:diff} are inspired by the recent work \cite{ildiz2024self} which characterizes the learnability of a ground-truth attention model via the next-token prediction objective in terms of the associated Markov transition matrix.


\section{Methodology: Selective Attention Layer} \label{sec:method}
Let us recap the self-attention mechanism in Transformer~\cite{vaswani2017attention}. Canonical softmax attention admits an input sequence $\X =[\x_1~\dots~\x_L]^\top\in\R^{L\times d}$ of length $L$ with embedding dimension $d$. We then project $\X$ to obtain key, query, and value embeddings ($\Kb=\X\W_k,~\Qb=\X\W_q,~\Vb=\X\W_v$) and compute the output of the dot-product attention as $Att(\Qb,\Kb,\Vb) = \sft{\frac{\Qb\Kb^\top}{\sqrt{d}}}\V$. Here $\sft{\cdot}:\R^L\rightarrow\R_+^L$ denotes the softmax nonlinearity that applies row-wise and $\W_q,\W_k,\W_v \in \mathbb{R}^{d \times d}$ are learnable weight matrices. In this paper, we mainly focus on casual language modeling where each token can only attend to previous tokens in the input.  

The uniform treatment of all tokens through the same softmax map could hinder the ability to control contextual \xuechen{sparsity} and relevance. For instance, it has been observed that current Transformer language models suffer from an attention dilution issue: the longer the input sequence, the flatter the attention distribution \cite{yao-etal-2021-self,chi2023attention}. A natural solution to the dispersed attention issue is to sharpen the self-attention distribution. Selective Attention aims to provide a general strategy to control \xuechen{spikiness} of the softmax adaptive to the query and value embedding, as well as the position of the token.





\begin{definition}[Selective Self-Attention (SSA)] Let $\X=[\x_1~\dots~\x_L]^\top\in\R^{L\times d}$ be an input sequence. Let $\tau_{k/q/v}(\cdot):\R^d\rightarrow\R^d$ be the inverse-temperature functions for keys, queries, and values, respectively. Then the embeddings for  keys ($\Kb$), queries ($\Qb$), and values ($ \Vb$) are computed as follows:
\[ 
\Kb=\tau_k(\X)\odot\X\W_k,~\Qb=\tau_q(\X)\odot\X\W_q,~\Vb=\tau_v(\X)\odot\X\W_v.
\] 
where $\odot$ denotes the elementwise product that assigns temperature to individual tokens. Selective Self-Attention (SSA) is then computed as $\sft{\frac{\Qb\Kb^\top}{\sqrt{d}}}\V$.
\end{definition}

In essence, SSA incorporates a temperature modulation mechanism into the attention framework to enhance selectivity and context control. The inverse-temperature function $\tau(\cdot)$ is data-dependent, allowing for dynamic adjustment of attention across different parts of the input sequence. In practice, we choose $\tau_{k/q/v}$ to be a scalar valued function as vector-valued temperature does not provide a significant advantage. It is also worth mentioning that we don't restrict $\tau_{k/q/v}$ to be non-negative. As a result, our temperature scaling strategy can be seen as an application of \emph{scalar gating} on K/Q/V embeddings, and hence, the SSA layer could also be referred to as \emph{Scalar-Gated Attention (SGA)} layer. The GitHub repo containing SSA implementation is provided in \url{https://github.com/umich-sota/selective_attention}. Below, we discuss the design choices underlying SSA.

\noindent$\bullet$ \textbf{Temperature scaling for query and value tokens.}  In an attention mechanism, the concepts of keys ($\Kb$), queries ($\Qb$), and values ($ \Vb$) play distinct roles in determining how information is weighted and combined across a sequence. Temperature functions can be applied to all of those components, designated as Key-temperature $\tau_{k}(\cdot)$, Query-temperature $\tau_{q}(\cdot)$ and Value-temperature $\tau_{v}(\cdot)$. We explore the advantages of each temperature function in in \Cref{sec:ablation}. In practice, we employ Query-temperature $\tau_{q}(\cdot)$ and Value-temperature $\tau_{v}(\cdot)$ but don't touch the original key embeddings. The query-temperature $\tau_{q}$ adjusts the spikiness of the attention map associated with the query according to its embedding and position in the context window. The value-temperature $\tau_{v}$ enhances the model’s ability to suppress irrelevant or noisy tokens, ensuring a refined aggregation of context window. In \Cref{sec:theory}, we provide insights into theoretical and empirical benefits of incorporating these terms.
While we keep the keys unmodified, guided by the intuition from word embeddings of~\cite{mikolov2013efficient} suggests that the similarity between a (key, query) pair should align with their cosine similarity. That is, $cos(key_1, query)>cos(key_2, query)$ should ideally imply that the $query$ attends more to $key_1$ compared to $key_2$. Assigning temperature/gating to scale the query vector does not change this order. However, if we assign distinct scalings to $key_1$ and $key_2$, we will end up with scenarios where attention scores are flipped i.e. $\tau_1*key_1^\top query < \tau_2*key_2^\top query$. In other words, our intuition is that assigning gating on keys will end up influencing their relative semantic similarities to queries (which could perhaps be better achieved via attention weights). This is in contrast to query-scaling which helps decouple the semantic similarity and contextual sparsity and the associated theoretical benefits (\Cref{query benefit sec} and \Cref{lem soft approx}).

\noindent$\bullet$ \textbf{Token-aware and position-aware temperature scaling.}
The data-dependent inverse-temperature function is composed of two distinct components $\tau(\x) = \tau^{tok}(\x)+\tau^{pos}(\x)$, $\x$ is a token within the sequence $\X$:  Token-aware Temperature Scaling $\tau^{tok}(\cdot)$ and Position-aware Temperature Scaling $\tau^{pos}(\cdot)$. Token-aware Temperature Scaling $\tau^{tok}(\cdot)$ is devised to modulate the influence of individual tokens within the sequence. The formula for this component is given by $\tau^{tok}(\x) = tanh (f(\x))$, where $f(\cdot)$ represents a trainable function that adjusts the impact of the token $\x$. The activation function $tanh(\cdot)$ is used to enable the scaling function to output both positive and negative temperatures; for instance, if we want to have the option to fully-suppress a token $\tau^{tok}(\x)$ can attain $\approx 0$. To address the issue of dispersed attention, where increasing length of the input sequence leads to a flatter attention distribution,  we introduce Position-aware Temperature Scaling. This is defined by $\tau^{pos}(\x) = 1 + \sigma(\alpha) log(n)$, where $n$ denotes the position of the token $\x$ within the sequence $\X=[\x_1~\dots~\x_L]^\top\in\R^{L\times d}$, $n\in[L]$. 
We remark that $n$ reflects the token length when computing the temperature of token $\x_n$ , aligning with our focus on causal attention where each token is restricted to attending only to previous tokens in the sequence. $\alpha$ is a parameter designed to modify the scale of the factor. The non-linearity $\sigma(\cdot)$ is the sigmoid function, employed to control the range of $\tau^{pos}$ and ensure the stability of the training process.

\noindent$\bullet$ \textbf{Weight sharing.} We introduce a weight sharing strategy to reduce the number of parameter overhead below 0.5\% (10x fewer) while maintaining the benefits of SSA.  Specifically, the Position-aware Temperature Scaling term, $\tau^{pos}(\x)$ only includes a single parameter $\alpha$, whereas the Token-aware Temperature Scaling term $\tau^{tok}(\x) = tanh (f(\x))$, relies on a trainable function $f(\cdot)$ defined as $\W_{tmp}\text{GeLU}(\W'_{tmp}\x)$,  involves separate trainable parameters $\mathbf{W}_{tmp}$ and $\mathbf{W}'_{tmp}$, which increases parameter load. To improve efficiency, we (re)use the attention weights $\W_{k/q/v}$ for the temperature module by setting $f(\x)=\W_{tmp}\text{GeLU}(\W_{k/q/v}\x)$. Here, SSA only adds the output layer of the MLP, a vector with few parameters. The approach only stores 3 vectors (not matrices) per attention head. This also have negligible inference/latency overhead because we don’t require additional matrix multiplication. These methods only require vector dot-products (at the output layer of the temperature module) and elementwise scaling of matrices. Other strategies can also be deployed to reduce the computational overhead. We describe feature-based approach which use simple token-level statistics, such as their frequencies in training corpus. Only constant parameters per head need to be stored that reduce the number of parameter overhead below 0.1\%. The deatils are shown in~\ref{app:weight-sharing}.


Finally, we discuss conceptual connections to \emph{sparse attention} methods in \Cref{app:connect}.

\section{Theoretical Insights into Selective Attention} \label{sec:theory}

Selective attention computes the query temperature based on the embedding and the position of the query. It also computes the value temperature based on the value embedding. In what follows, we discuss how these three components provably enhance expressivity of the attention mechanism. 
\subsection{The benefits of incorporating query embedding}\label{query benefit sec}

\noindent\textbf{Decoupling semantics from specificity.} Consider two words: ``Hinton'' and ``Scientist''. The former is a specific instance of the latter. As a result, while we expect token embeddings of these two words to have high cosine similarity, they might benefit from different attention maps. Specifically, ``Hinton'' refers to a specific person and we expect it to have a more targeted attention to the context associated with it. 
We argue that query-temperature can aid optimization by retaining semantic similarity while allowing for distinct \emph{specificity}. More formally, by specificity we are referring to the contextual \xuechen{sparsity} level of a query. Denoting the combined key-query weights to be $\W=\W_q\W_k^\top$ as a problem-agnostic measure of specificity, we will consider the magnitude of the query embedding. That is, given query token $\qb$, define $\text{spec}_{\W}(\qb):=\tn{\W^\top\qb}$. It is well-established \cite{tarzanagh2023maxmargin} that in order for attention map to be more sparse (hence higher specificity), the norm of the query embedding, or more generally the operator norm of $\W$, has to grow larger, justifying this definition. The following Lemma shows that, without TS, the attention weights within softmax have to be lower bounded by the ratio of $\emph{specificity difference}$ to ${\emph{semantic distance}}$.

\begin{lemma}\label{vanilla lower bound} Let $\W=\W_q\W_k^\top\in\R^{d\times d}$ be the combined query-key matrix. Let $\ab,\bb\in\R^d$ be unit norm token embeddings associated with the \emph{specific} and \emph{general} token respectively. Suppose we wish to achieve specificities $\text{spec}_{\W}(\ab)\geq L_a$ and $\text{spec}_{\W}(\bb)\leq L_b$. Then, the associated $\W$ obeys $\|\W\| \geq \frac{L_a-L_b}{\tn{\ab-\bb}}$.
\end{lemma}
Above $L_a-L_b$ is the \text{specificity\_difference} whereas $\tn{\ab-\bb}$ is the semantic distance. The proof follows from the triangle inequality $\|\W\|\geq \frac{\tn{\W^\top(\ab-\bb)}}{\tn{\ab-\bb}}\geq \frac{\tn{\W^\top\ab}-\tn{\W^\top\bb}}{\tn{\ab-\bb}}\geq \frac{L_a-L_b}{\tn{\ab-\bb}}$. 

\noindent \textbf{Comparison to Selective Attention.} In SSA, the effective attention weight matrix for a query $\qb$ is $\W=\tau(\qb)\cdot \W_q\W_k^\top$. To achieve the same specificity in Lemma \ref{vanilla lower bound} with SSA, we can set the temperatures as $\tau(\ab)=L_a$, $\tau(\bb)=L_b$, and KQ-weights as $\|\W\|=1$ (e.g.~via $\W=\Iden_d$). This achieves the desired specificities while maintaining that \emph{effective weights} are upper bounded as $\|\W_{\ab}\|,\|\W_{\bb}\|\leq \max(L_a,L_b)$. In other words, the required norm growth is entirely decoupled from the semantic distance between the queries.

In essence, this highlights that without query-selectivity, the model weights have to grow excessively to assign different specificity to similar words. In practice, this is expected to create performance bottlenecks: (1) As the weights grow, optimization may slow down along certain directions due to vanishing softmax derivative and, (2) even if the optimization is successful, the final model could overfit or be overly sensitive to small perturbations in the context, hindering test accuracy.
\begin{wrapfigure}{r}{0.55\textwidth}
\centering
\includegraphics[width=0.45\textwidth]{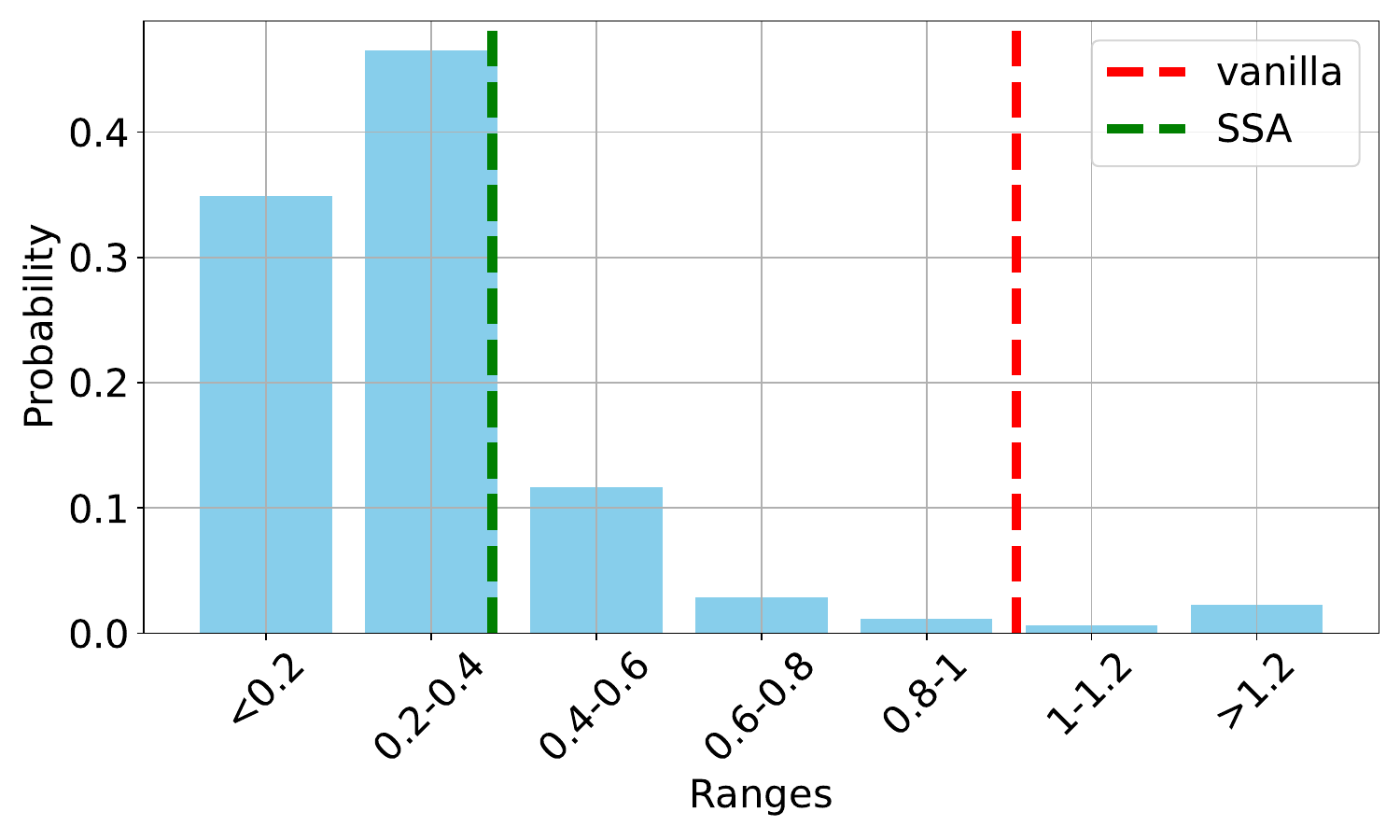}
\vspace{-8pt}
\caption{The operator norm of $\W$ with and without Query-temperature scaling, scaled by $\times 10^{3}$. The figure depicts the distribution across 1000 tokens. The dashed line is the average norm. Notably, the norm of the vanilla attention layer is approximately three times larger than that of SSA(dashed red line compare to green line). Furthermore, the vanilla attention layer exhibits a lower \xuechen{spikiness} score (0.39) compared to SSA (0.26), where a lower value indicates higher \xuechen{spikiness}.}
\vspace{-20pt}
\label{tab:norm}
\end{wrapfigure}
 
This is also verified by our experiments. To study the norm growth of attention weights, we train Pythia from scratch, trainig with the SlimPajama dataset~\cite{shen2023slimpajama}(our pre-training setting) and evaluate on Wikitext dataset. We examine the average norm of combined query-key matrix weight $\|\W\|$ from the average of all layers within the model. Additionally, we quantify the \xuechen{spikiness} of the attention map computed as the ratio of the $l_{1}$–norm to the
squared $l_{2}$–norm and normalized by the length, defined as$\frac{\|\boldsymbol{s}\|_{1}}{\|\boldsymbol{s}\|^{2}L}$, $\boldsymbol{s}$ where $\boldsymbol{s}$ is the softmax probability vector. It takes values from 0 to 1. A smaller value indicates a sparser vector. We compute the average of the first 1000 tokens of the Wikitext dataset. The results shown in \Cref{tab:norm} align with the theory. The attention weights for selective attention are smaller than the original ones, while the attention is sparser.
\begin{figure}[htbp]
\centering
\begin{subfigure}[b]{0.2\textwidth}
\includegraphics[scale=0.2]{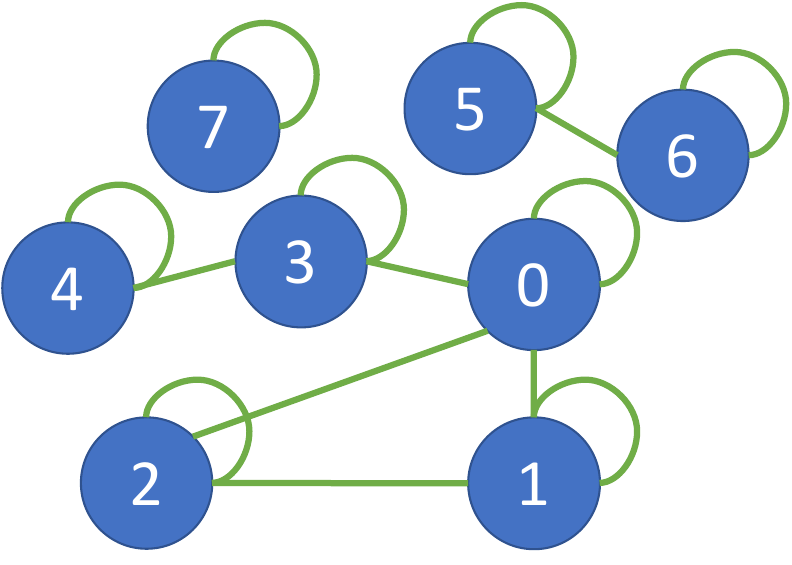}
  \caption{The graph.}
\end{subfigure}\label{fig:graph_figure}
\begin{subfigure}[b]{0.23\textwidth}
\includegraphics[scale=0.15]{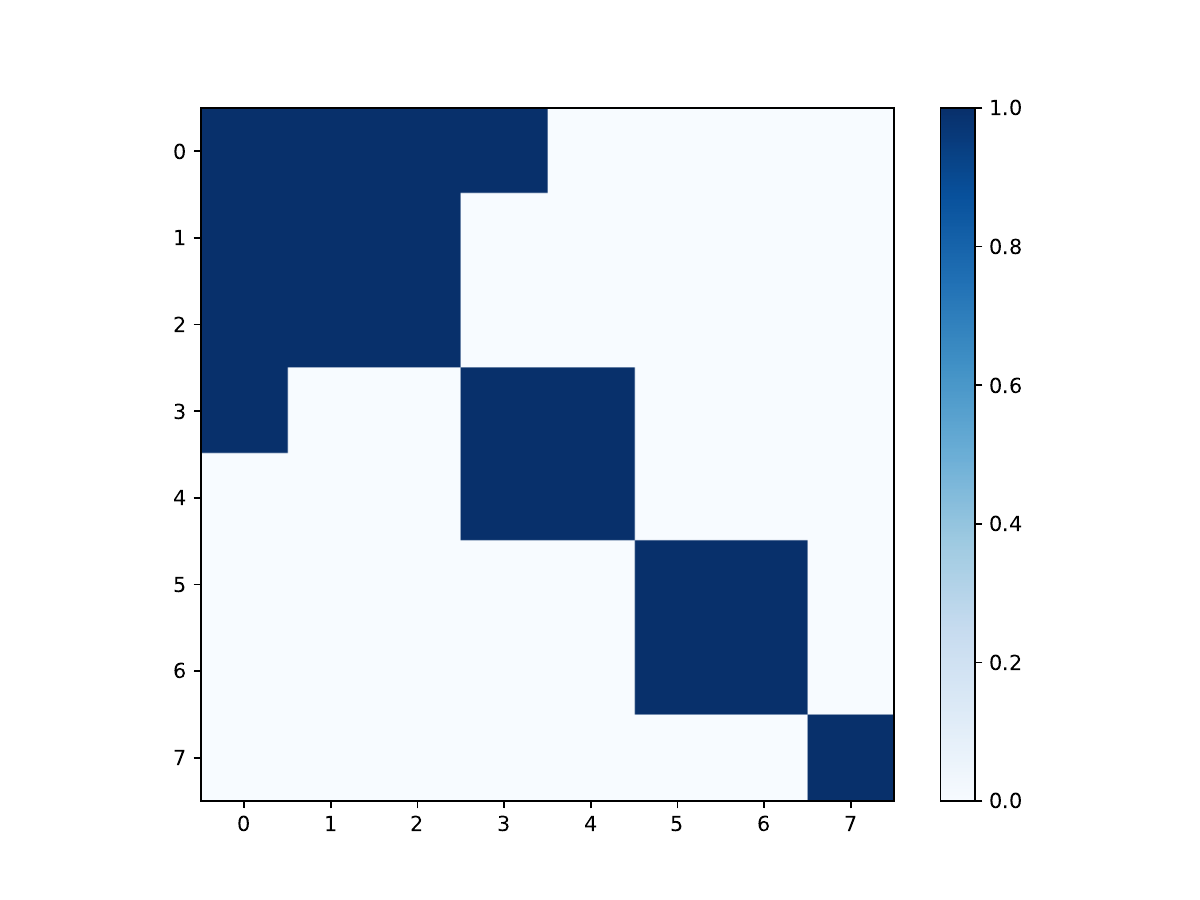}
\vspace{-5pt}
  \caption{Ground-truth \\token transitions $\Pb_\star$}
\end{subfigure}
\begin{subfigure}[b]{0.23\textwidth}
\includegraphics[scale=0.15]{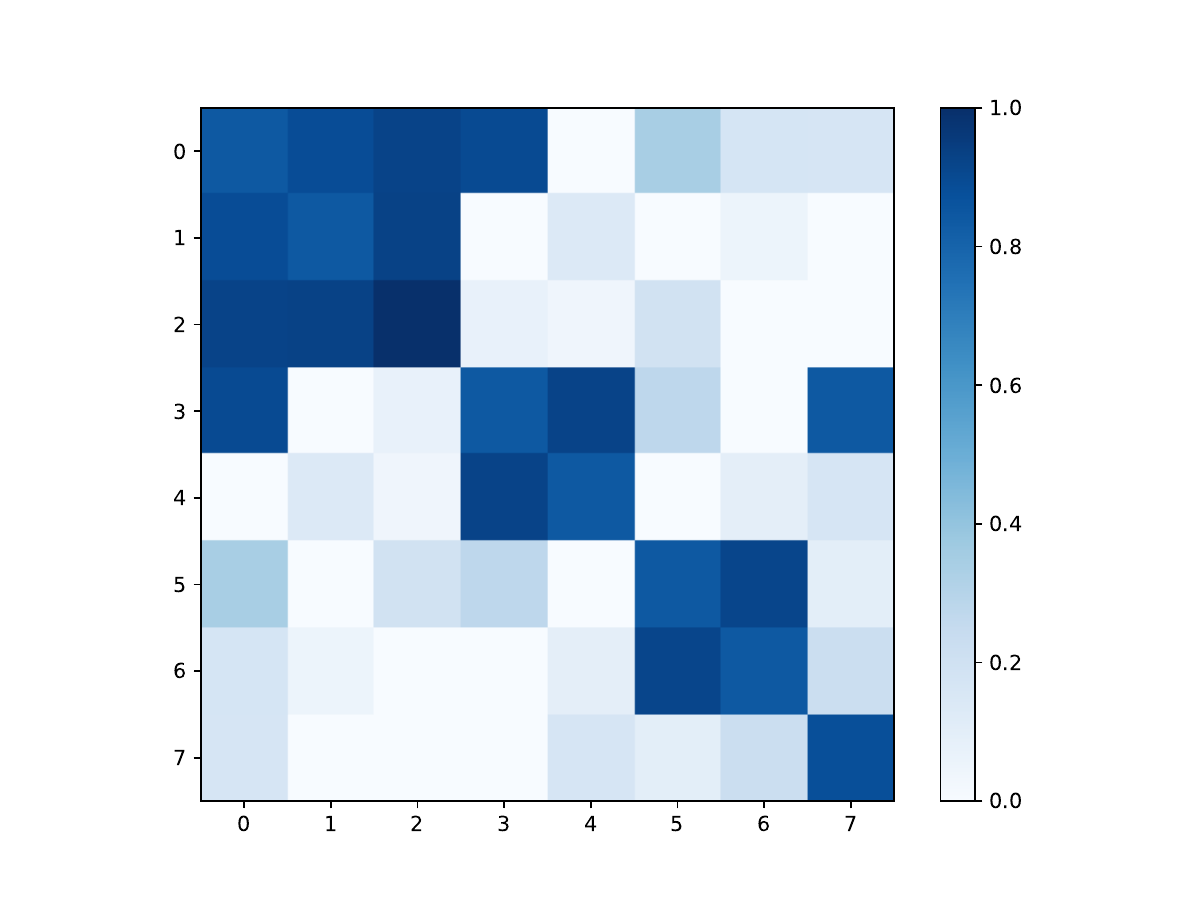}
\vspace{-5pt}
  \caption{$\hat{\Pb}$ learned \\by SSA\\}
\end{subfigure}
\begin{subfigure}[b]{0.23\textwidth}
\includegraphics[scale=0.15]{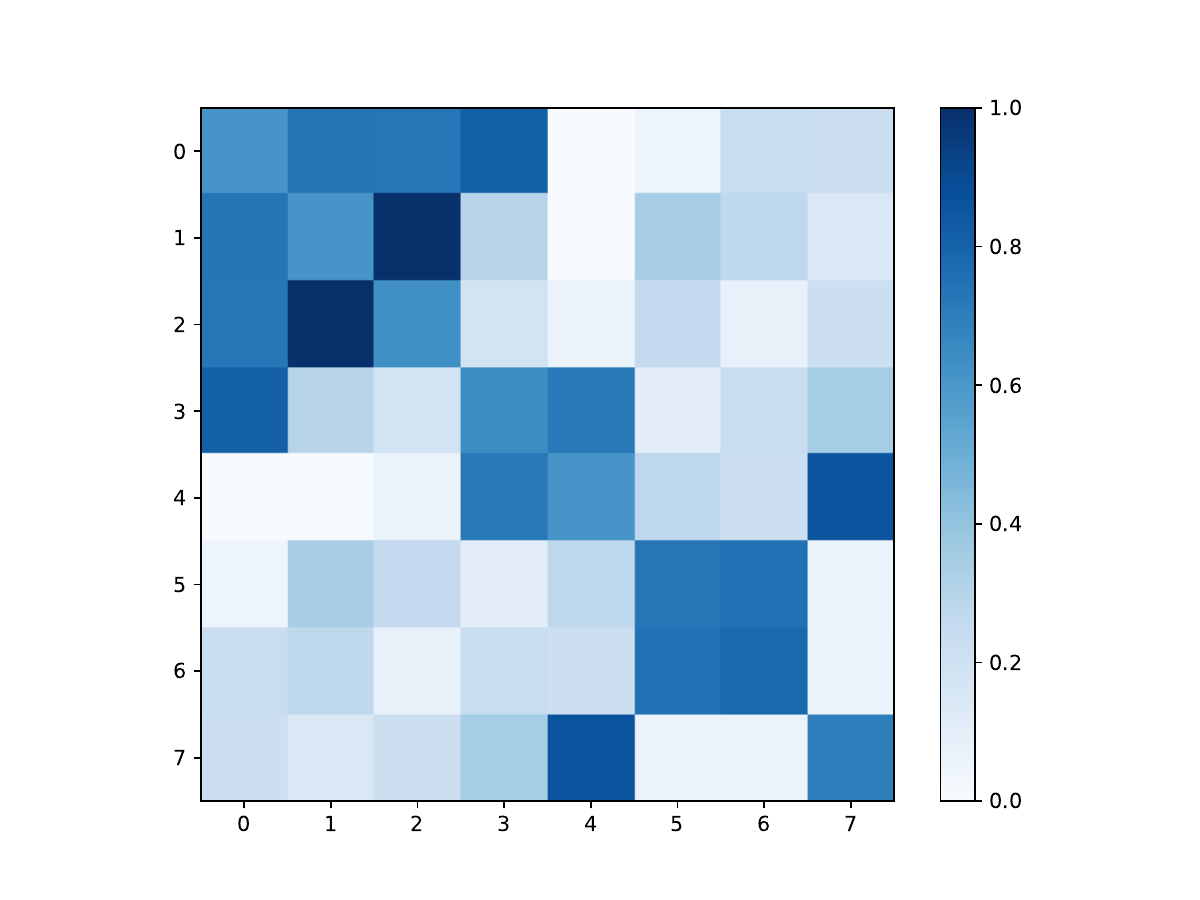}
\vspace{-5pt}
  \caption{$\hat{\Pb}$ learned by\\ Self-Attention}
\end{subfigure}
\caption{We compare 1-layer SSA and 1-layer attention when solving next-token prediction on a small vocabulary of size 8. (a) is the graph associated to the token transition dynamics. (b) is the the pairwise token transition matrix of this vocabulary. Each row of $\Pb_\star$ represents an attention map where a particular token is the query and all tokens in the vocabulary serve as keys (see Sec \ref{query benefit sec} for details). The transition matrix $\hat{\Pb}$ estimated by SSA in (c) is sharper and more closely resembles the optimal $\Pb_\star$. SSA achieves a smaller cross-entropy loss compared to vanilla attention, 0.009 vs 0.0126. The $\ell_1$ approximation error of the attention map of SSA is also smaller than that of vanilla attention, 0.358 vs 0.543. }
\label{fig:diff}
\end{figure}
\noindent\textbf{Expressivity benefits of query-selectivity.} A closely related consideration is whether query-selectivity can enhance expressivity.
We expect that through query-temperature, the same attention head will have an easier time expressing sparse and dense attention maps associated with distinct queries. To formalize this, we investigate the ability of a single (selective) attention head to express a target attention map between all tokens in a discrete vocabulary. Let $\Vc={\eb_i}_{i=1}^K$ be a vocabulary of $K$ tokens. To capture all $K^2$ pairwise interactions of these tokens, we first form the sequence $\Eb=[\eb_1~\dots~\eb_K]^\top\in\R^{K\times d}$ where each token appears uniquely and then study $K$ attention maps associated with individual queries, i.e., $\att{\Eb,\eb_i}$ for $1\leq i\leq K$. Stacking these together as rows, we study the $K\times K$ attention matrix $\att{\Eb}$. For standard attention with weights $\W$, this is given by $\att{\Eb,\W}=\sft{\Eb\W\Eb^\top}$, whereas for query-selective attention, $\att{\Eb,\W}=\sft{\tau(\Eb)\odot \Eb\W\Eb^\top}$.

Thanks to the softmax nonlinearity, $\att{\Eb}$ is a stochastic matrix where rows add up to 1. This matrix can be viewed as a Markov chain transition between different tokens, which motivates a fundamental question: \emph{Can query-selective attention help express a larger class of stochastic matrices?} Intuitively, we expect that if a stochastic matrix $\Pb_\star$, which we wish to express via $\att{\Eb}$, exhibits a lot of \xuechen{spikiness} variation across its rows (i.e., different queries), selectivity can better capture these.

This can be verified with a token generation experiments. Recall that we expect ``bacteria'' to attend to more words compared to ``salmonella''. 
We might expect more general words to have a larger number of neighbors in a graph.
Accordingly, we abstract the vocabulary, which comprises words with various levels of specificity, into a simple undirected graph. This is depicted in \Cref{fig:graph_figure}. Additionally, the stochastic matrix $\Pb_\star$ can be derived from this graph, with the results displayed in \Cref{fig:diff}(a). To build the estimation of the stochastic matrix $\Pb_\star$ training, we conduct next token prediction experiments.

\noindent \emph{Token generation setting:} Let $X\in\Vc^L$ be a sequence of length $L$ drawn from $\Vc$. Suppose $X$ ends with $q:=x_L$. The token $Y=x_{L+1}$ that follows $X$ will be drawn uniformly from $q$ or one of the neighbors of $q$. This neighborhood is parameterized via the latent attention map $\Pb_\star$ which will govern the generation process. Let $\Eb=[\eb_1~\dots~\eb_N]^\top$ be the token embeddings associated with the vocabulary $\Vc$. Assume elements of $\Eb$ have unit $\ell_2$ norm. In data generation, we simple sample input sequences containing each token in the vocabulary precisely once, and sample the next token according to the attention map $\Pb_\star$, that is, the row of $\Pb_\star$ that corresponds to the final query token. We then fit a one-layer self-attention or SSA model $f(\X)$ to approximate this latent dynamics. Concretely, we predict the next token $\hat{Y}$ of $f(\X)$ according to the distribution $g(\X)=\sft{\Cb f(\X)}\in\R^N$. Here $\Cb\in\R^{N\times d}$ is the linear prediction head. As loss measure on how well we fit to the latent $\Pb_\star$ dynamics, use the cross entropy distance between $g(\X)$ and the true label $Y$. Through this, we wish to formalize and visualize the intuitions on why ``salmonella'' deserves a lower temperature than ``bacteria''. Further experimental details are described in \Cref{app:detail}. 

In our experiments, besides smaller cross-entropy loss, we find that Selective Attention achieves a better approximation of $\Pb_\star$ as shown in \Cref{fig:diff}. To evaluate the similarity between the attention map $\Pb_\star$ and $\hat{\Pb}$, we also define the $\ell_1$ distance between the attention maps, namely,
\[ 
\texttt{err\_map}=\|\hat{\Pb}-\Pb_\star\|_1.
\] 
We find that the $\texttt{err\_map}_\text{SSA}$ is also much lower than $\texttt{err\_map}_\text{vanilla}$ (0.358 vs 0.543). Additionally, SSA naturally assigns lower temperatures to tokens with fewer neighbors. This is in line with our expectations as fewer neighbors imply a sparser attention map. The results are shown in \Cref{tab:graph_neighbor}.

\begin{table}[t]
    \centering
    \caption{Temperature for each depth. Nodes with the same \# of neighbors share the same temperature.}
     \label{tab:graph_neighbor}
    \begin{tabular}{c|c|c|c|c}
\hline
 \# of neighbors(including itself) & 1 & 2    & 3 & 4\\ \hline
 nodes index & 7 & 4,5,6    & 1,2,3 & 0 \\ \hline
 temperature &  0.002   & 0.019 & 0.152    & 0.751\\ \hline
    \end{tabular}
        \vspace{-10pt}
\end{table}
To further formalize this, we revisit Lemma \ref{vanilla lower bound} in terms of softmax map. Let $K=2$ and $\Pb_\st=\begin{bmatrix}1-\gamma&\gamma\\0&1\end{bmatrix}$ be the target pairwise attention map. Here second token is highly specific (only selects itself) whereas the first token is less specific when $0<\gamma<1$. The following proposition establishes a variation of Lemma \ref{vanilla lower bound} when approximating $\Pb_\st$.


\begin{proposition}\label{lem soft approx}
Suppose the embeddings $\eb_1, \eb_2$ have unit $\ell_2$ norm with correlation $\rho = \eb_1^\top \eb_2$. Fix $0 < \eps \leq \frac{1}{2} \min(\gamma, 1-\gamma)$ and $\Gamma = \left| \log \left( \frac{1-\gamma}{\gamma} \right) \right|$. For any $\W$ obeying $\|\Pb_\st - \sft{\Eb \W \Eb^\top}\|_{\infty} \leq \eps$, we have that $\|\W\| \geq \frac{\|\eb_1 - \eb_2\|^{-1}}{\sqrt{2 - 2\rho^2}} \left( \log \left( \frac{1}{4\eps} \right) - \Gamma \right)$. Conversely, Selective Attention can achieve this $\eps$-approximation with weights bounded as $\tau(\eb_{1,2}) \cdot \|\W\| \leq \|\eb_1 - \eb_2\|^{-1} \max \left( \log \left( \frac{1}{\eps} \right), \frac{\Gamma}{\sqrt{1-\rho^2}} \right)$.
\end{proposition}






\subsection{The benefits of incorporating query position}

The need for position-dependent scaling arises from the fact that, for a fixed weight matrix $\W=\W_q\W_k^\top$, the attention scores $\s^L=\sft{\X\W^\top\x_L}$ become diluted as sequence length $L$ grows. Specifically, for retrieval-type tasks, the model may want to concentrate softmax scores $\s^L$ on a single token. However, assuming unit norm tokens, the top probability in $\s^L$ is upper bounded via $\tin{\s^L}\leq \frac{1}{1+(L-1)e^{-2\|\W\|}}$. This implies that, to enforce $\tin{\s^L}$ to be constant, we require the spectral norm lower growth rate of $\|\W\|\geq 0.5\log L+O(1)$. This motivates our logarithmic scaling strategy which was also proposed by \cite{peng2023yarn,chi2023attention}.

Here we provide a more formal justification on the optimal temperature scaling rule by describing a simple yet insightful task which is not solvable by a single attention head unless temperature scaling is employed. Specifically, we consider a setting where the sequence exhibits \emph{feature imbalances} where frequent tokens start dominating the context and potentially overwhelm the less frequent but relevant tokens. 

\noindent\textbf{Imbalanced token setup:} Suppose the input sequence $\X=[\x_1~\dots~\x_L]^\top$ is composed of a minority token $\ab\in\R^d$ and a majority token $\bb\in\R^d$, that is, $\x_i\in\{\ab,\bb\}$ for all $i\in[L]$. For each position, we will simply ask the model to output a target mixture of $\ab$ and $\bb$, namely, $\y=\alpha\ab+(1-\alpha)\bb$ for some $\alpha\in(0,1)$. Thus, using a 1-layer causal attention, we study the following objective by calculating the loss between target $\y$ and each attention output:
\begin{align}
\Lc(\W)=\frac{1}{L}\sum_{n=n_0}^L\tn{\y-\X^\top\csf{n}{\tau_n\cdot \X\W^\top\x_n}}^2.\label{imba risk}
\end{align}
Above, $\tau_n$ is the inverse-temperature for the $n^{th}$ position. Here, $n_0$ is a burn-in period to simplify our exposition: $n_0$ is the smallest number such that both $\ab$ and $\bb$ appear at least once within the first $n_0$ tokens\footnote{This is without loss of generality, otherwise, causal attention would output a fixed vector (either $\ab$ or $\bb$) regardless of the attention weights.}. Additionally, let $n_a$ be the number of tokens $\x_i$ that are equal to $\ab$ within $i\in[n]$. We have the following theorem. 
\begin{proposition} \label{position thm}Assume $\ab,\bb$ are unit Euclidean norm and linearly independent. Define the imbalance ratio $\kappa_n=(n-n_a)/n_a$ for $n\in[L]$. There is a $\W_\st$ such that, setting $\tau_n=\log\kappa_n+\log\frac{\alpha}{1-\alpha}$, $\Lc(\W_\st)$ minimizes the risk \eqref{imba risk} to achieve $\Lc(\W_\st)=0$.
    
Conversely, consider the problem instance with \emph{target mixture} of $\alpha=1/2$, \emph{second-quadrant imbalance} of $2 \geq \kappa_n\geq 1$ for $L/4\leq n\leq L/2$ and \emph{fourth-quadrant imbalance} of $\kappa_n\geq 4$ for $n\geq 3L/4$. If we employ flat temperature $\tau_n=1$ for all $n\in[L]$, for any choice of attention weights $\W\in\R^{d\times d}$, we have the lower bound $\Lc(\W)>1/500$.
\end{proposition}

\Cref{position thm} inspired our design of position-aware temperature scaling. Intuitively, as $n$ increases, the sequence may include less related tokens, leading to an increase in $\kappa_n$. When $\kappa_n$ follows power-law $\kappa_n=n^\pow$, we recover the logarithmic temperature scaling rule of $\tau_n=\texttt{const}+\pow\cdot \log n$.  Consequently, our Position-aware Temperature Scaling function $\tau_n$ is designed as $\tau^{pos}(\x) = 1 + \sigma(\alpha) log(n)$, $n$ is the position length, $\alpha$ is the trainable parameter, $\sigma$ is the  non-linearity function sigmoid. The function is motivated by, other paper's rules \cite{peng2023yarn,menon2020long,li2021autobalance,zhang2024class}.


\subsection{The benefits of incorporating value embedding}


Within attention, value embeddings ($\Vb$) are transformed using only a linear projection. Consequently, each token's contribution to the output is a weighted sum based on the attention scores, with these weights adjusted linearly. In sequences with many tokens, irrelevant or noisy tokens can negatively influence the attention mechanism. Because value embeddings are linearly projected, they may not be able to fully distinguish between relevant and irrelevant tokens. The value-temperature scaling acts as a nonlinear scalar weighting function. By adjusting the temperature, we aim to control the impact of each token, suppressing the influence of irrelevant or noisy tokens. This helps emphasize more relevant tokens, thereby improving the quality of the context representation. We motivate the potential benefits of TS on value embeddings through the following synthetic denoising task.

\textbf{Denoising task} \label{value temp def} Let $[K]$ be the token alphabet with embeddings $(\eb_i)_{i=1}^K$. Assume $d=K$ and $\eb_i$'s are standard basis. Consider the following data distribution $(\X,\y)\sim\Dc$ where $\X=[\x_1~\dots~\x_L]^\top\in\R^{L\times d}$ is the input sequence and $\y\in\R^d$ is the target label.
\begin{itemize}
    \item Draw $q\sim\texttt{Unif}([K])$. Set $\y=\eb_q$. 
    \item Let $(\z_i)_{i=1}^L$ be IID noise vectors with $\Nn(0,\sigma^2\Iden)$
    \item $\x_L=\eb_q+\z_L$. For $i\in[L-1]$, $\x_i$ is determined by a Bernoulli distribution with a parameter of $\alpha$, selecting between $\eb_q+\z_i$ and $\z_i$. Consequently, $\alpha$ of the tokens are signal tokens $\eb_q+\z_i$.
\end{itemize}
The denoising objective is minimizing the MSE risk 
\[ 
\Lc(f) = \E_{\Dc}[\tn{\y-\text{norm}(\hat{\y})}^2]
\]
where 
$\text{norm}(\hat{\y}) = \hat{\y}/\tn{\hat{\y}}$, $\hat{\y}$ is the output of model $f(\cdot)$, $\hat{\y} = f(\X) $.

To solve this task, the attention model $f(\X)$ should intelligently combine the tokens within $\X$ to approximate the denoised target $\eb_q$. Importantly, the model will strictly benefit from eliminating the pure noise tokens, i.e., instances with $\x_i=\z_i$. Note that the value projection of the attention matrix will not suffice to denoise the input sequence. The reason is that $q$ is uniform, and signal tokens span the whole space. Thus, we will benefit from a nonlinear denoising procedure.

To test this intuition, we use a 1-layer single-head attention model, denoted as different $f(\cdot)$ to minimize the denoising objective. We compare the model with value-selectivity to the following baselines:


\begin{enumerate}
\item \emph{Vanilla Attention:} The standard 1-layer single-head attention model, $\hat{\y}_{att}=\text{Att}(\X)$
\item \emph{Value-selective self-attention:} 1-layer Selective Self-Attention (SSA). $\hat{\y}_{SSA}=\text{SSA}(\X)$. Since this is a synthetic task, as a proxy for the token-aware temperature scaling, we use the selection function $\max_{j\in[d]} x_{ij}\geq 1/2$. Intuitively, when noise $\sigma\lesssim 1/\sqrt{\log d}$, thresholding with the largest entry will detect the signal tokens.
\item \emph{Naive averaging:} Directly average the tokens, $\hat{\y}_{naive}=\frac{1}{L}\sum_{i=1}^L \x_i$.
\item \emph{Bayes optimal estimator:} $\hat{\y}_{opt}=\frac{1}{|S|}\sum_{i\in S} \x_i$ where $S\subset[L]$ is the ground-truth set of signal tokens distributed as $\eb_q+\z_i$.
\end{enumerate}

\begin{table}[t]
    \centering
    \caption{We apply normalization to attention output and compute the MSE risk. }
    \begin{tabular}{c|c|c|c}
\hline
Vanilla & Value-selective  & Naive averaging &Bayes optimal estimator\\ \hline
1.390 & 0.071 & 2.058 &0.003\\ \hline
    \end{tabular} \vspace{-10pt}
        \label{tab:MSE_risk}
\end{table}
The resulting MSE risks are displayed in \Cref{tab:MSE_risk}. We set $d = k = 8$ and $\alpha = \frac{1}{4}$. With the addition of the value-selection function, the model achieved a loss comparable to the optimal estimator, indicating successful suppression of noisy tokens. In contrast, while vanilla softmax self-attention performs similarly to naive averaging, it fails to sufficiently denoise, resulting in a much larger loss compared to our value-selective attention.
\section{Empirical Evaluations}\label{sec:exp}
\subsection{Standard Benchmarks}
Drawing on theoretical insights, we assess the performance of SSA on NLP tasks by integrating SSA into established models such as GPT-2~\cite{radford2019language}, Pythia~\cite{biderman2023pythia}, Llama~\cite{touvron2023llama} and Llama3~\cite{dubey2024llama}. 
\begin{table}[t]
\centering
\small
\caption{Experiment results for model pretraining and finetuning. For perplexity (ppl), lower is better, and for accuracy (acc), higher is better. }
\label{tab:my_label}
    \centering  
    \resizebox{1.05\textwidth}{!}{%
\begin{tabular}{@{}l|ccc|ccccccc|c@{}}
\toprule
\thead{Model} & {\thead{Wikitext \\ ppl$\downarrow$}} & {\thead{Lambada\_std \\ ppl$\downarrow$}} & {\thead{Lambada\_openai \\ ppl$\downarrow$}} &
{\thead{Lambada\_std \\ acc$\uparrow$}} & {\thead{Lambada\_openai \\ acc$\uparrow$}} &{\thead{Piqa \\ acc$\uparrow$}} & {\thead{Hella \\ acc\_norm$\uparrow$}} & {\thead{Winogrande \\ acc$\uparrow$}} &
{\thead{Arc-E \\ acc$\uparrow$}} & {\thead{Arc-C \\ acc\_norm$\uparrow$}} & {\thead{Average \\ acc$\uparrow$}} \\ \midrule
\multicolumn{11}{c}{Finetune} \\ \midrule
GPT2 & 36.503 & 51.631  & 29.134 & 0.340 & 0.451 & 0.584& 0.313& 0.476& 0.457& 0.221& 0.406\\
 +SSA base & 34.618 &  50.412 & 27.235 & 0.361 & 0.469  & 0.610& 0.338& 0.512& 0.479& 0.249& 0.431 \\
 +SSA weight sharing& 35.147 & 50.832 & 27.905&  0.357&  0.465& 0.603& 0.334& 0.500& 0.472& 0.243 & 0.425 \\\midrule
Pythia-160m & 26.681&47.996&24.102&0.383&0.494  & 0.674& 0.362& 0.542& 0.503& 0.277 & 0.462\\
 +SSA base & 26.514 & 47.945 & 23.956 & 0.388 & 0.513  & 0.688& 0.375& 0.557& 0.530& 0.291& 0.477 \\
 +SSA weight sharing& 26.780 &  47.961& 24.027&  0.386& 0.509 & 0.685& 0.369& 0.553& 0.524& 0.285 & 0.473 \\\midrule
Pythia-410m & 20.310 &  42.694& 21.895&  0.418&  0.542&0.696 & 0.372& 0.547& 0.561& 0.288 & 0.489\\
 +SSA base &  19.976& 42.689 & 21.704&  0.430& 0.553 &0.714 & 0.381& 0.558& 0.572& 0.302 & 0.501 \\
 +SSA weight sharing&  20.190&  42.692& 21.810&  0.428& 0.549 & 0.707& 0.380& 0.551&0.566 & 0.295 & 0.497\\\midrule
Llama & 19.764 &  28.023 & 16.513 & 0.426 & 0.574  & 0.704&  0.377 &0.549 & 0.595& 0.302 & 0.504\\
 +SSA base& 19.305 & 27.627 & 15.860 & 0.428 & 0.581 & 0.710&  0.388 & 0.562&0.618 & 0.336 & 0.518 \\
 +SSA weight sharing& 19.512 & 27.892 & 16.038 & 0.426 & 0.579  & 0.708& 0.385& 0.557& 0.608& 0.331 & 0.513 \\\midrule
Llama3-8b &12.416&24.002&13.954&0.481&0.684& 0.772& 0.544&0.698& 0.780&0.463 & 0.632\\
 +SSA base & 10.982 & 23.671 & 12.052&0.489 & 0.690& 0.779& 0.550& 0.703&0.787 & 0.472 & 0.639 \\
 +SSA weight sharing& 11.498 & 23.805 & 10.164& 0.487 & 0.687 & 0.776&0.548 &0.701 &0.784 & 0.471 & 0.636\\
\midrule
\multicolumn{11}{c}{Pretrain} \\ \midrule
GPT2 & 35.813 & 104.225 &  42.187 & 0.216 & 0.304 & 0.608& 0.309& 0.462& 0.359& 0.186 & 0.349\\
 +SSA base & 33.528 & 103.933  & 40.960 & 0.221 &  0.318 & 0.631& 0.317& 0.480&0.365 & 0.203& 0.362 \\
 +SSA weight sharing  & 34.601 & 104.004 & 41.326 & 0.219 & 0.312 & 0.622& 0.312& 0.469& 0.365& 0.197 & 0.356 \\\midrule
Pythia-160m   & 27.943 & 75.487  & 34.406 & 0.279 & 0.351 & 0.630&0.348&0.498&0.401&0.219 & 0.389\\
 +SSA base        & 26.912 & 72.891  & 33.126 & 0.294 & 0.360 & 0.661&0.359 & 0.508&0.426 & 0.230 & 0.405\\
 +SSA weight sharing & 27.046&73.071&33.814&0.291&0.360&  0.660 & 0.352  &0.503  &0.421 & 0.221  & 0.401\\\midrule
Pythia-410m &22.516&69.814&32.781&0.321&0.371& 0.655&0.357 &0.530 &0.441 & 0.234& 0.416\\
 +SSA base &21.402&68.553&31.269&0.336&0.387&0.660 & 0.363& 0.536&0.449 & 0.237& 0.424\\
 +SSA weight sharing  &21.980&69.041&31.458&0.331&0.384&0.658 &0.362 & 0.534& 0.445& 0.237& 0.422 \\
\bottomrule
\end{tabular}%
}
\end{table}
Our methodology includes both pre-training and fine-tuning to evaluate SSA's performance and efficiency. For the pre-training evaluation, we train the model from scratch on the SlimPajama dataset~\cite{shen2023slimpajama}. Subsequently, we evaluate the model on various downstream zero-shot tasks, including Wikitext~\cite{merity2016pointer}, Lambada~\cite{paperno2016lambada}, Piqa~\cite{bisk2020piqa}, Hella~\cite{zellers2019hellaswag}, Winogrande~\cite{sakaguchi2021winogrande}, Arc-E, and Arc-C~\cite{clark2018think}.
This approach is widely used for measuring the performance and generalization capabilities of pretrained large language models across diverse tasks~\cite{arora2024simple,biderman2023pythia,gu2023mamba}.
For the fine-tuning evaluation, we start by loading the official pre-trained model and then fine-tune it on the downstream tasks. Unlike pre-training, where the downstream tasks are unseen during training, fine-tuning involves direct training on the tasks. This allows the model to better approximate the token distribution and understand the text domain. Details of the models are provided in \Cref{app:detail}. 

Our primary results are shown in \Cref{tab:my_label}. Based on the theoretical insights and ablation study results, we conduct both Token-aware and Position-aware Temperature Scaling on query $\Qb$, and value $\Vb$. We observe that across various models and datasets, incorporating SSA consistently enhances performance. Notably, experiments with larger and more recent models, such as Llama3-8B and Pythia 410M, confirm that SSA improves accuracy across across model scales and architectures. We further introduce a weight sharing strategy that reduces the number of parameter overhead to less than 0.5\% while preserving the benefits of SSA and still outperforming the standard transformer. This underscores the value of selectivity irrespective of its precise implementation. Thus, our improvements are not arising from an increase in the parameter count, but rather from the strategic integration of SSA. Additionally, we have also explored a \emph{feature-based} method to further enhance SSA’s parameter efficiency. In a nutshell, rather than training an MLP, we select the temperature as a function of \emph{token-level features}, such as the frequency of a token in the training corpus, by fitting a single scalar parameter. This process requires only O(1) additional weights (<0.01\% of total). Further details and results are provided in \Cref{app:weight-sharing}.

For the ablation study, we fine-tuned the models on the Wikitext dataset to compare the influence of each component, using the same dataset and training configurations as those in the real experiments. The results are shown in \Cref{sec:ablation}. Among the results, we observe that deploying both Token-aware and Position-aware Temperature Scaling on $\Qb$ and $\Vb$ independently could achieve significant improvement, aligning with our theoretical insights. Additionally, combining Key and Query temperatures can achieve additional improvement. Moreover, between token-aware and position-aware temperature scaling, the latter demonstrates a more consistent improvement across different scenarios, while combining them can achieve the best overall result. We also compare with more baselines including~\cite{menon2020long,li2021autobalance,zhang2024class} and the results are shown in \Cref{app:add_exp}. Our method consistently outperforms the baselines. 
\begin{wrapfigure}{r}{0.4\textwidth}
 \vspace{-0pt}
\centering
\includegraphics[width=0.25\textwidth]{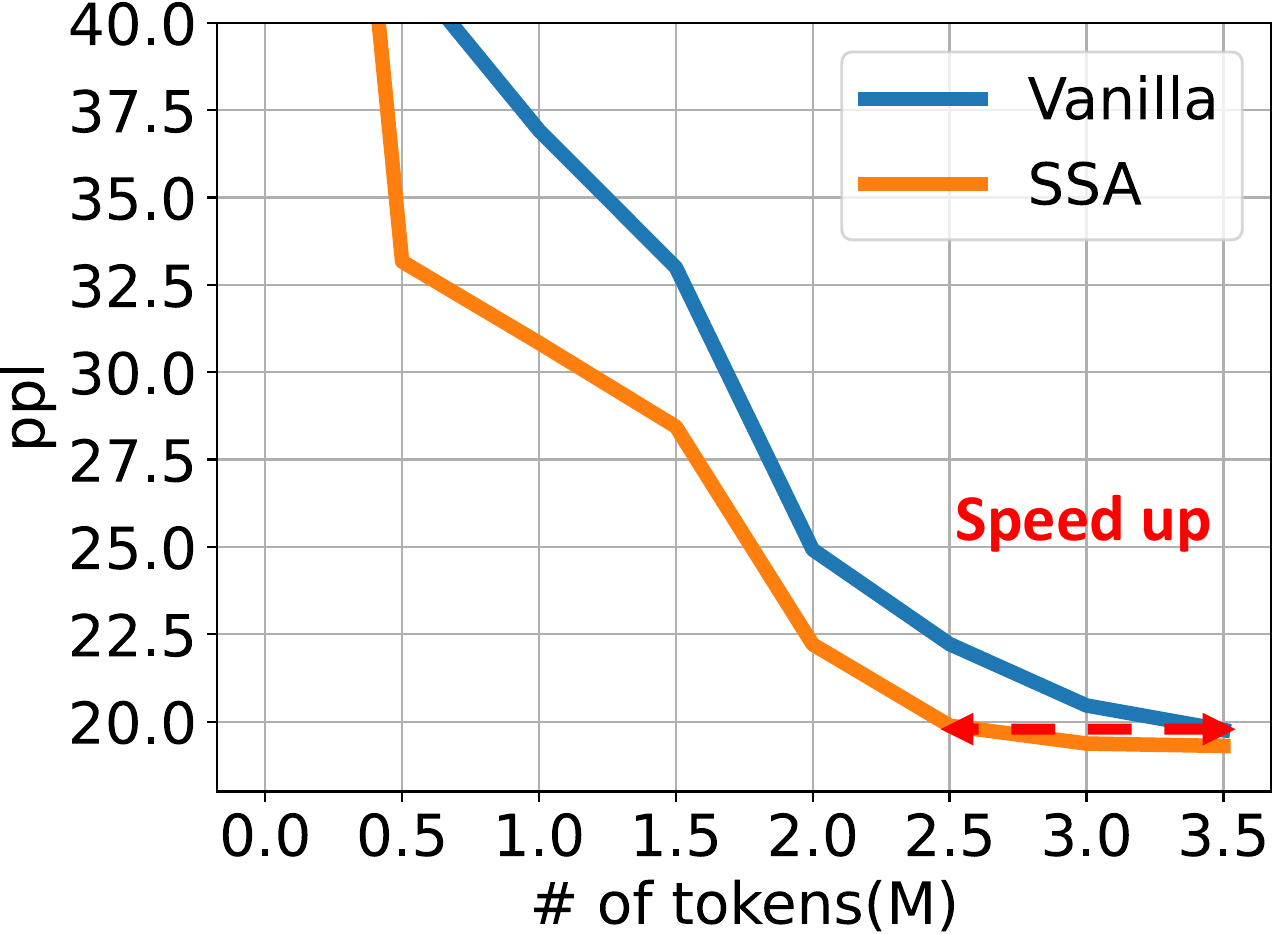}
\vspace{-0pt}
\caption{Comparison of training curves. SSA provides reasonable benefits in terms of training speedup.} \label{fig:acc}
\vspace{-10pt}
\end{wrapfigure}

Additionally, SSA can accelerate the training process by achieving comparable performance with fewer tokens. This efficiency not only reduces the demand on computational resources but also shortens the time required to effectively train models. We illustrate this efficiency by plotting the training results when fine-tuning the Llama model on the Wikitext dataset, both with vanilla attention layer or SSA, in \Cref{fig:acc}. The results indicate that SSA can accelerate training, achieving similar performance with 1.45× reduction in pretraining steps. 
\subsection{Passkey Retrieval}
We also examines the perfromance on the passkey retrieval task as defined in \cite{peng2023yarn,mohtashami2023landmark}.This is a synthetic task to measure a model’s ability to retrieve a simple passkey (i.e., a five-digit number) within a large amount of otherwise meaningless text. We performed 10 iterations of the passkey retrieval task with the passkey placed at a random location uniformly distributed across the evaluation context window. Intuitively, SSA could better solve this task by assigning different token-level temperatures to digits vs words. For our evaluation of the fine-tuned Pythia, SSA leads to substantial improvement (from 56.9\% to 74.4\%), as seen in \Cref{tab:passkey}. 
\begin{table}[h!]
\centering
\small
\vspace{-10pt}
\caption{Passkey retrieval performance of various models.}
\label{tab:passkey}
\begin{tabular}{c|cccc}
\toprule
\thead{Model} & {Original} & {+SSA}  &
{+SSA(weight sharing)}\\ \midrule
Pythia-160m & 56.89& 74.41&  66.90\\
Llama& 77.62 & 89.53 &  89.45 \\
\bottomrule
\end{tabular}
\vspace{-10pt}
\end{table}

\vspace{-10pt}\section{Conclusions, Limitations, and Future Directions}\label{sec discuss}
\vspace{-5pt}
We have introduced the Selective Self-Attention layer, which augments the softmax nonlinearity with a principled temperature-scaling strategy. SSA shows consistent benefits and augments the performance of existing transformer-based models such as Pythia and Llama 2. We also provide theoretical insights into the benefits of query, value, and positional selectivity.

\noindent\textbf{Future research.} Based on SSA, there are several interesting research avenues to pursue. Firstly, our method can extend to linear attention strategies. While we can use the same method for value embeddings, for queries, we can train an additive bias term on attention similarities rather than using temperature scaling. Secondly, based on the visual benefits of SSA on Figure \ref{fig:diff}, it would be interesting to explore how SSA can help the interpretability and quality of the attention maps. Overall, SSA has the potential to assist in more principled use of transformers in language, vision, and other modalities.


\noindent\textbf{Limitations.} Our work focuses on the canonical softmax-attention mechanism, which suffers from the quadratic computation bottleneck. As mentioned above, extending our method to linear attention can mitigate computational costs. Another direction to enhance efficiency is building stronger connections to sparsity and understanding how SSA can benefit and be integrated with sparse attention algorithms.

\subsection*{Acknowledgements}

This work was supported in part by the National Science Foundation grants CCF-2046816, CCF-2403075, CCF-2008020, the Office of Naval Research award N000142412289, an Adobe Data Science Research award, and gifts by Open Philanthropy and Google Research. 



\bibliography{refs}
\bibliographystyle{plain}
\newpage 
\appendix

\section{Implementation details} \label{app:detail}
For the next token prediction experiment that substantiates the expressivity benefit of query-selectivity, as detailed in \Cref{fig:diff} and \Cref{tab:graph_neighbor}, we employ the Adam optimizer to train a model. This model consists of a single-layer, single-head attention mechanism, accompanied by a tokenizer and a fully connected layer. The tokenizer embeds the discrete sequence to continuous embedding $\Eb$. The fully connected layer is used as the classifier to predict the node index. We set the learning rate at $1e^{-4}$. The training loss is the cross-entropy loss. In our experiments, $L = N = 8$. For SSA, we implement Token-aware Temperature Scaling for the query matrix $\Qb$. We assign a scaling parameter to each group of nodes that share the same number of neighbors. To have better visualization, we do normalization to plot the attention map $\Pb_\star$ and $\hat{\Pb}$. For the experiments shown in \Cref{tab:MSE_risk}, we also use the Adam optimizer and learning rate $1e^{-4}$. But the objective is the MSE risk.

For our empirical evaluation, we utilize several models. We employ GPT-2, which has 124 million parameters, and use the official OpenAI GPT-2 checkpoints that were pre-trained on the WebText dataset~\cite{radford2019language} for our finetuning experiments. For Pythia, our experiments are conducted with a model size of 160 million and 410 million parameters, using the official checkpoint pre-trained on the Pile dataset~\cite{gao2020pile}. Lastly, for Llama, we utilize the smallest variant available, with 7 billion parameters, and similarly fine-tune using the official pre-trained model. As the training configuration, we train with 3.5 million tokens for fine-tuning and 15B tokens for pre-training. We always use the AdamW optimizer~\cite{loshchilov2017decoupled}, $\beta_1 = 0.9$ and $\beta_2 = 0.95$. We set learning rate $1e^{-6}$ with no weight decay and no warmup. The pre-training takes about 2 hours using 4 A40 and fine-tuning takes about 2 days. We use FlashAttention~\cite{dao2023flashattention} to accelerate the training. For weight sharing, each head shares the same funtion. All the experiments are conducted with 4 or 8 A40 or L40S. We can directly reuse FlashAttention\cite{dao2022flashattention}, which significantly improves model efficiency.
\section{Additional experiments}
\subsection{Ablation study} \label{sec:ablation}
For the ablation study, we conducted fine-tuning on the Wikitext dataset to compare the influence of each component, using the same dataset and training configurations as those in the real experiments. The models are evaluated with perplexity (ppl). 
\subsubsection{Key-temperature, Query-temperature, and Value-temperature}
To evaluate the benefits of applying temperature scaling to $\Kb$, $\Qb$, and $\Vb$, we conducted an ablation study, examining each component individually and in combination. For a fair comparison, both position-aware and token-aware temperature scaling were applied to all components. The results, detailed in \cref{tab:ablation_kqv}, indicate that modifying $\Qb$, and $\Vb$ independently yields clear benefits, whereas alterations to $\Kb$ result in performance that is similar to, or even worse than, the baseline vanilla attention layer. The results align well with the theoretical analysis presented in \cref{sec:theory}. However, when $\Qb$ and $\Vb$ are combined, we observe consistent improvements. These findings led us to develop our final algorithm, which applies temperature scaling to both $\Qb$ and $\Vb$.

\begin{table}[t!]
\centering
\caption{Fine-tuning experiment results for language models on the Wikitext dataset, showcasing baseline and variations with different components ($\Qb, \Kb, \Vb$). }
\label{tab:ablation_kqv}
\begin{tabular}{c|cc}
\toprule
Configuration & Pythia  & GPT2  \\
\midrule
Baseline      & 28.781       & 36.503     \\
$\Qb$            & 27.416       & 34.832     \\
$\Kb$           & 28.715       & 36.443     \\
$\Vb$            & 27.980       & 35.857     \\
$\Qb$, $\Vb$         & 26.514       & 34.618     \\
$\Kb$, $\Qb$, $\Vb$       & 26.603       & 34.609     \\
\bottomrule
\end{tabular}
\end{table}
\subsubsection{Token-aware Temperature Scaling, Position-aware Temperature Scaling}
We also conduct experiments to investigate the benefits of Token-aware and position-aware Temperature Scaling applied to Query $\qb$ and value $\vb$. The results are shown in \cref{tab:pos_tok}. Token-aware Temperature Scaling positively impacts both Query $\qb$ and value $\vb$, whereas Position-aware Temperature Scaling shows smaller improvement on value $\vb$, aligning with our theoretical insights. Furthermore, when compared to GPT-2, Pythia—which features a more advanced positional encoding\cite{su2024roformer} —demonstrates fewer improvements. This suggests that while new strategies may mitigate the dispersed attention issue, our Selective Self-Attention (SSA) method still offers additional improvements.
\begin{table}[t!]
\caption{Investigate the benefits of Token-aware Temperature Scaling, Position-aware Temperature Scaling.} \label{tab:pos_tok}
\centering
\begin{tabular}{c|c|ccc|ccc}
\hline
\multirow{2}{*}{model} & \multirow{2}{*}{vanilla} & \multicolumn{3}{c|}{$\qb$}                           & \multicolumn{3}{c}{$\vb$}                           \\ \cline{3-8} 
                      &                          & \multicolumn{1}{c|}{$\tau^{pos}+\tau^{tok}$} & \multicolumn{1}{c|}{$\tau^{pos}$} & $\tau^{tok}$ & \multicolumn{1}{c|}{$\tau^{pos}+\tau^{tok}$} & \multicolumn{1}{c|}{$\tau^{pos}$} & $\tau^{tok}$ \\ \hline
 Pythia                  &          28.781                & \multicolumn{1}{c|}{27.416} & \multicolumn{1}{c|}{27.995} & 27.503 & \multicolumn{1}{c|}{27.980} & \multicolumn{1}{c|}{28.342} & 27.975 \\ \hline
GPT2                     &          36.503                & \multicolumn{1}{c|}{34.832} & \multicolumn{1}{c|}{34.970} & 35.064 & \multicolumn{1}{c|}{35.857} & \multicolumn{1}{c|}{36.320} &  35.617\\ \hline
\end{tabular}
\end{table}

\subsection{Parameter-efficient SSA: Weight sharing and featurization}\label{app:weight-sharing}
Here, we also introduce a feature-based approach to improve parameter efficiency. In a nutshell, rather than training an MLP, we select the temperature as a function of \emph{token-level features}, such as the frequency of a token in the training corpus, by fitting a single scalar parameter. This process requires only O(1) additional weights per attention head (<0.01\% of total). This is inspired by the logit adjustment strategy of~\cite{menon2020long} which sets the cross-entropy temperature as a function of class frequencies. 

Our evaluations on feature-based SSA are provided in \Cref{tab:app_weight}. We find that, while the feature-based method is beneficial and highly parameter-efficient, it can be sensitive to feature selection and exhibits more variability across datasets. 


\begin{table}[t]

\caption{Comparing different SSA parameterizations}
\centering
\small
\begin{tabular}{c|ccccc}
\hline
\thead{Model} & {\thead{Wikitext \\ ppl$\downarrow$}} & {\thead{Lambada\_std \\ ppl$\downarrow$}} & {\thead{Lambada\_openai \\ ppl$\downarrow$}} &
{\thead{Lambada\_std \\ acc$\uparrow$}} & {\thead{Lambada\_openai \\ acc$\uparrow$}}\\
\midrule
\multicolumn{6}{c}{Finetune} \\ \midrule
Pythia & 26.681&47.996&24.102&0.383&0.494  \\
Pythia +SSA base & 26.514 & 47.945 & 23.956 & 0.388 & 0.513 \\
Pythia +SSA weight sharing& 26.780 &  47.961& 24.027&  0.386& 0.509\\
Pythia + SSA feature-based &27.048&47.966&24.114&0.387&0.499\\
\midrule
\multicolumn{6}{c}{Pretrain} \\ \midrule
Pythia   & 27.943 & 75.487  & 34.406 & 0.279 & 0.351 \\
Pythia +SSA base        & 26.912 & 72.891  & 33.126 & 0.294 & 0.360 \\
Pythia +SSA weight sharing & 27.046&73.071&33.814&0.291&0.360 \\
Pythia +SSA feature-based & 27.281&73.614&33.794&0.287&0.357 \\
\bottomrule
\end{tabular}
\label{tab:app_weight}
\end{table}

\subsection{Ablation of Different Parameterizations} \label{app:add_exp}

In addition to the functions we propose for temperature design, we also explore alternative approaches. Instead of employing our Position-aware Temperature Scaling function, we use a constant parameter, as suggested in other studies \cite{menon2020long,li2021autobalance,zhang2024class}. We also compare with the temperature scaling method proposed by \cite{peng2023yarn}. Furthermore, in place of our Token-aware Temperature Scaling, we adopt a simpler approach by directly utilizing token frequency and training only a scale parameter. These experiments were conducted using the Pythia model, fine-tuned on the Wikitext dataset. The outcomes of these comparative analyses are presented in \Cref{tab:func}. Among those baselines, we consistently outperform their results.

\begin{table}[t!]
\centering
\caption{Conducting different functions.}
\label{tab:func}
\begin{tabular}{c|c|c}
\toprule
\multicolumn{2}{c|}{  Configuration }& Pythia    \\
\midrule
\multicolumn{2}{c|}{  Vanilla}      & 28.781           \\ \hline
\multirow{2}{*}{Token} &Yarn \cite{peng2023yarn} &  27.602   \\
& Constant       & 28.058          \\ \hline
Position & Frequency       & 27.360    \\  \hline
Position+Token & SSA         & 26.514            \\
\bottomrule
\end{tabular}
\vspace{-15pt}
\end{table}

\section{Proofs}
\subsection{Proof of \Cref{lem soft approx}}\label{app:proof_lemma2}
\begin{proof}
Given embeddings $\eb_1, \eb_2$ with unit $\ell_2$ norm, their correlation is $\rho = \eb_1^\top \eb_2$. 

First, consider the approximation error bound $\|\Pb_\st - \sft{\Eb \W \Eb^\top}\|_{\infty} \leq \eps$. To achieve this, the weight matrix $\W$ must satisfy the inequality.

To derive a lower bound on $\|\W\|$, observe that:
\begin{align*}
\|\Pb_\st - \sft{\Eb \W \Eb^\top}\|_{\infty} &\leq \eps \\
\|\Pb_\st - \sft{\Eb \W \Eb^\top}\|_{\infty} &= \left\| \frac{1}{1+e^{-\Eb \W \Eb^\top}} - \Pb_\st \right\|_{\infty} \\
&\leq \eps \\
\intertext{Using the fact that $\eb_1$ and $\eb_2$ are unit vectors and $\rho = \eb_1^\top \eb_2$, we have:}
\|\Eb \W \Eb^\top\|_{\infty} &\geq \frac{1}{4\eps} - \Gamma.
\end{align*}

Now, the norm $\|\W\|$ is given by:
\begin{align*}
\|\W\| &\geq \frac{\|\eb_1 - \eb_2\|^{-1}}{\sqrt{2 - 2\rho^2}} \left( \log \left( \frac{1}{4\eps} \right) - \Gamma \right).
\end{align*}

Conversely, to achieve $\eps$-approximation using Selective Attention, the weights need to be bounded such that:
\begin{align*}
\tau(\eb_{1,2}) \cdot \|\W\| &\leq \|\eb_1 - \eb_2\|^{-1} \max \left( \log \left( \frac{1}{\eps} \right), \frac{\Gamma}{\sqrt{1 - \rho^2}} \right).
\end{align*}

Therefore, the selective attention avoids the $1/\sqrt{1-\rho^2}$ dependence on the $\log(1/\eps)$ term, decoupling the high-specificity requirement (small $\eps$) from the semantic similarity of the tokens. 

This completes the proof of \Cref{lem soft approx}.
\end{proof}

\subsection{Proof of Proposition \ref{position thm}}
\begin{proof} We first show the success direction. Set $\W$ such that $\bb^\top\W=0$ and $\ab^\top\W\ab=\ab^\top\W\bb=1$. Such $\W$ exists thanks to the linear independence of $\ab,\bb$. Now plugging this $\W$ into \eqref{imba risk}, for each position, regardless of whether $\x_n=\ab$ or $\x_n=\bb$, we obtain
\[ 
\X^\top\csf{n}{\tau_n\cdot \X\W\x_n}=\frac{n_ae^{\tau_n}}{n_ae^{\tau_n}+(n-n_a)}\ab+\frac{(n-n_a)e^{\tau_n}}{n_ae^{\tau_n}+(n-n_a)}\bb.
\] 
We wish to ensure
\[ 
\frac{n_ae^{\tau_n}}{n_ae^{\tau_n}+(n-n_a)}=\frac{1}{1+(n/n_a-1)e^{-\tau_n}}=\frac{1}{1+\kappa_ne^{-\tau_n}}=\alpha.
\] 
This in turn implies 
\[ 
\kappa_ne^{-\tau_n}=\frac{1-\alpha}{\alpha}\iff \tau_n=\log\kappa_n+\log\frac{\alpha}{1-\alpha}.
\]

Next, we discuss the failure case of flat temperature. To do so, we will lower bound the loss over the queries $\x_n=\bb$. Set $M=e^{(\bb-\ab)^\top\W\bb}$. Following same argument as above, for fixed temperature, $\W$ will output a non-adaptive composition of the form 
\[ 
\X^\top\csf{n}{\X\W\bb}=\frac{1}{1+M\kappa_n}\ab+\frac{M\kappa_n}{1+M\kappa_n}\bb.
\]
Thus, the loss function will be lower bounded by (accounting for the prediction error in $\ab,\bb$ terms and their orthogonality)
\[ 
\Lc(\W)\geq \min_{M>0}\sum_{n:\x_n=\bb} (0.5-\frac{1}{1+M\kappa_n})^2+(0.5-\frac{M\kappa_n}{1+M\kappa_n})^2=\min_{M>0}\sum_{n:\x_n=\bb} 2\cdot(0.5-\frac{1}{1+M\kappa_n})^2.
\] 
Now since $\kappa_n\geq 1$ over both second, at least $1/2$ of the queries are $\x_n=\bb$. Similarly, at least $4/5$ of the queries are $\x_n=\bb$ over the last quadrant. We will lower bound the loss over the two scenarios depending on $M\geq 1/3$ or not. 

First, suppose $M\geq 1/3$, in that case, using $\kappa_n\geq 4$ over the last quadrant, the loss is lower bounded by
\[ 
\Lc(\W)\geq \min_{M>0}\frac{1}{N}\sum_{n\geq 3N/4,\x_n=\bb}2\cdot(0.5-\frac{1}{1+M\kappa_n})^2\geq \frac{2}{5}(0.5-\frac{1}{1+4/3})^2> 0.002.
\] 
where we used the fact that there are $\geq \frac{N}{5}=\frac{N}{4}\cdot\frac{4}{5}$ queries with $\x_n=\bb$ over the last quadrant.

Similarly, suppose $M\leq1/3$, in that case, using $\kappa_n\leq 2$ over second quadrant, the loss is lower bounded by
\[ 
\Lc(\W)\geq \min_{M>0}\frac{1}{N}\sum_{N/2\geq n\geq N/4,\x_n=\bb}2\cdot(0.5-\frac{1}{1+M\kappa_n})^2\geq \frac{1}{4}(0.5-\frac{1}{1+2/3})^2= 0.0025.
\] 
where we used the fact that there are at least $N/8$ queries with $\x_n=\bb$ over the second quadrant. Combining these two cases, we found that, for any choice of $\W$, the loss is lower bounded by $0.002$.
\end{proof}

\section{Further Discussion on the Sparsity and Temperature Connection}\label{app:connect}

\noindent\textbf{Connections between sparsity and temperature.} 

To formally study the sparsity and temperature connection, let us consider a fixed attention row $n$ and introduce:
\begin{itemize}
\item $\s(\tau)=\csf{n}{\tau\cdot\X\W\x_n}$, the scaled attention scores with inverse-temperature $\tau>0$.
\item $\bs(\kappa)=\csf{n}{\X\W\x_n,\kappa}$ denote the sparse attention scores where the top-$\kappa n$ entries are retained and the rest are set to $0$ where $0\leq \kappa\leq 1$.
\end{itemize}

The connection between sparsity and temperature scaling is clear. For instance, the top entry of $\s(\tau)$ will be decreasing in $\tau$ whereas the entropy of $\s(\tau)$ will be increasing. Here, we would like to establish how temperature scaling rule can be mapped to a sparsity rule. We will do this under a power-law relevance assumption on the attention scores. Here, we assume that the attention scores admit two values and the fraction of larger/relevant attention scores follow a power-law as context window grows.
\begin{assumption}[Power-law relevance] Consider the vector of raw attention scores $\ab=\X\W\x_n$. Each entry of $\ab$ is either $c$ or $c_+=c+\gamma$ for some $\gamma>0$. Additionally, $n^{-\pow}$ fraction of the entries are equal to $c_+$ for some $\pow>0$.
\end{assumption}
Above $c_+$ is the score attained by the salient tokens, $\gamma$ is the score advantage of salient tokens over rest of the tokens, and $\pow$ dictates the fraction of the salient tokens. To proceed, we have the following lemma which identifies condition under which TS and sparse-attention exhibit the same softmax temperature behavior.
\begin{lemma} For any choice of $\tau=1+\alpha\log n>0$ and corresponding sparsity $\kappa=\frac{n^{-\alpha\gamma}}{1-n^{-\pow}}+n^{-\pow}$, we have that $\tin{\s(\tau)}=\tin{\bs(\kappa)}$.
\end{lemma}
This reveals the clear connection between temperature scaling and sparsification rules. Simplifying the above, this lemma advocates that the sparsification rule should follow the power law decay of $\kappa\approx n^{-\alpha\gamma\wedge \pow}$. Consistent with this lemma, our experiments demonstrate that sparsification with power-law results in respectable performance.


\begin{proof} We first compute the top entry of $\s(\tau)$ as follows 
\[ 
\frac{s_1^\tau}{\sum_{i=1}^ns_i^{\tau}}=\frac{e^{\gamma\tau}}{ n^{1-\pow}e^{\gamma\tau}+(n-n^{1-\pow})}=\frac{1}{n^{1-\pow}+n (1-n^{-\pow})e^{-\gamma\tau}}
\] 
We similarly compute the top sparse attention score as
\[ 
\frac{s_1}{\sum_{i=1}^{\kappa}s_i}=\frac{e^{\gamma}}{ n^{1-\pow}e^{\gamma}+(\kappa-n^{-\pow})n}= \frac{1}{n^{1-\pow}+(\kappa-n^{-\pow}) e^{-\gamma}n}.
\] 
Combining these, the top softmax probabilities are matched by setting 
\[ 
(\kappa-n^{-\pow}) e^{-\gamma}= (1-n^{-\pow})e^{-\gamma\tau}\iff \kappa=\frac{e^{-\gamma(\tau-1)}}{1-n^{-\pow}}+n^{-\pow}.
\] 
\end{proof}
We need to clarify that our method is not about sparse approximation of the attention map and instead aims to control the “spikiness of attention”. The “spikiness of attention” can be viewed as an “effective sparsity” which can be quantified through $L_\infty$ norm, $L_1/L_2$ ratio, or inverse-entropy of the softmax map. This discussion will also better clarify what is meant by “contextual sparsity” throughout the paper and distinguish it from (hard) sparsity targeted in \cite{pagliardini2023fast,ren2023sparse}.

\section{Theoretical Considerations}





\subsection{Hierarchical vocabulary}\label{app:Hierarchical}
\noindent \textbf{Hierarchical vocabulary.} Consider a $k$-ary tree of depth $D$: Each node has exactly $k$ children, except at depth $d$. Such a tree has $1+k+k^2+\dots+k^{D}=N=(k^{D+1}-1)/(k-1)$ nodes. The tree will correspond to the words/tokens in the vocabulary $\Vc$ of size $N$. 

\noindent\textbf{Token generation rule:} Let $X\in\Vc^L$ be a sequence of length $L$ drawn from $\Vc$. Suppose $X$ ends with $q:=x_L$. The token $Y=x_{L+1}$ that follows $X$ will be drawn from $q$ or the children of $q$ available in the context window. If $q$ is at depth $l$, it can attend to a total of $(k^{D+1-l}-1)/(k-1)$ unique tokens, including itself. Let $\Dc_{XY}$ denote the data distribution $(Y,X)$ where $Y$ is drawn uniformly from one of the \emph{child tokens of $x_L$} available in the context window $X$.

The claims below aim to formalize the benefits of SSA for modeling the hierarchical token generation process. Let $\Eb=[\eb_1~\dots~\eb_N]^\top$ be the token embeddings associated to the vocabulary $\Vc$. Assume elements of $\Eb$ are unit $\ell_2$ norm. During training, we embed the discrete sequence $X$ into $\X=[\eb_{x_1}~\dots~\eb_{x_N}]$.

\begin{claim}[Benefits on attention map]\label{claim1} Consider the attention map $\map(\X)=\sft{\tau(\x_L)\cdot\X\W\x_L}$. Note that $\map$ is a function of $\W,\Eb,\tau$. Define the ideal attention map $X$ to be $\map^\st(X)$ which uniformly attends to the children of $x_L$ and assigns zero probability to other tokens. Define the population error
\[ 
\texttt{err\_map}(\Eb,\W,\tau)=\E_{X\sim\Dc_X}\left[\|\map(\X)-\map^\st(X)\|_1\right].
\] 
Under suitable assumptions (see remark below), QSSA is provably better than vanilla self-attention i.e.~having $\tau(\x)$ improves attention capability by reducing $\texttt{err\_map}(\Eb,\W,\tau)$.
\end{claim}

\begin{claim}[Benefits on prediction] Let $f(\X)$ be an attention layer (SSA or vanilla). Suppose we sample the next token $\hat{Y}$ from $f(\X)$ according to the distribution $g(\X)=\sft{\Cb f(\X)}\in\R^N$. Here $\Cb\in\R^{N\times d}$ is the linear prediction head. As loss measure, use the expected total-variation (TV) distance between $g(\X)$ and the true label $Y$. Under suitable assumptions, $g(\X)$ with QSSA $f(\X)$ fits better to the hierarchical distribution compared to vanilla $f(\X)$.
\end{claim}

\end{document}